\documentclass[11pt]{article}

\usepackage[]{acl}

\usepackage{times}
\usepackage{latexsym}

\usepackage{CJKutf8}

\usepackage[T1]{fontenc}
\usepackage[utf8]{inputenc}
\usepackage{microtype}
\usepackage{inconsolata}
\usepackage{graphicx}
\usepackage[size=tiny]{todonotes}
\usepackage{tikz}
\usepackage{amsmath}
\usepackage{booktabs}
\usepackage{svg}

\def\checkmark{\tikz\fill[scale=0.4](0,.35) -- (.25,0) -- (1,.7) -- (.25,.15) -- cycle;} 

\definecolor{darkgreen}{rgb}{0.0, 0.5, 0.0} 
\newcommand{\green}[1]{\textcolor{darkgreen}{#1}}
\newcommand{\red}[1]{\textcolor{red}{#1}}
\newcommand{\h}[1]{\textcolor{black}{#1}}
\newcommand{\rev}[1]{\textcolor{black}{#1}}
\newcommand{\revv}[1]{\textcolor{black}{#1}}
\newcommand{\f}[1]{\textcolor{black}{#1}}

\newcommand{\dataset}[0]{\textit{ExCAM40k}}

\newcommand{\se}[1]{\textcolor{black}{#1}}
\newcommand{\reb}[1]{\textcolor{black}{#1}}

\title{%
  \makebox[\textwidth][c]{%
    \raisebox{-0.42\height}{\includegraphics[width=1.6em]{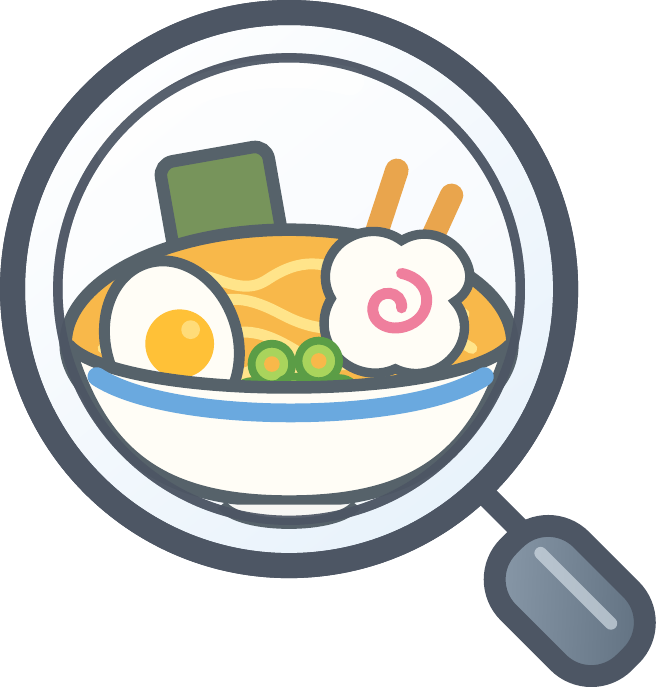}}%
    \hspace{0.6em}%
    \parbox[t]{0.67\textwidth}{\centering\bfseries ExCAM: Explainable Cultural Awareness Metrics}%
  }%
}

\author{
  \textbf{Christoph Leiter\textsuperscript{1,2}},
  \textbf{Haiyue Song\textsuperscript{3}},
  \textbf{Hour Kaing\textsuperscript{3}},
  \textbf{Jin Tei\textsuperscript{3}},
\\
  \textbf{Hideki Tanaka\textsuperscript{3}},
  \textbf{Masao Utiyama\textsuperscript{3}},
  \textbf{Steffen Eger\textsuperscript{1,2}}
\\
  \textsuperscript{1}University of Mannheim, Germany,\\
  \textsuperscript{2}University of Technology Nuremberg, Germany,\\
  \textsuperscript{3}National Institute of Information and Communications Technology
\\
  \small{
    \textbf{Correspondence:} \href{mailto:email@domain}{christoph.leiter@uni-mannheim.de}
  }
}

\begin{document}
\maketitle
\begin{abstract}
Evaluating the cultural awareness of large language models is crucial to ensure the fairness of generated text and the generalizability of applications across the world. Recent benchmarks explore cultural goods like food or values like behavior in stressful situations through the lens of question answering or text generation tasks. However, creating these benchmarks requires time-intensive and costly human annotations. Also, benchmarks that evaluate cultural awareness in free text are scarce and often rely on dated evaluation mechanisms. To address this gap, we introduce ExCAM, an \textbf{Ex}plainable \textbf{C}ultural \textbf{A}wareness \textbf{M}etric, \revv{which is}, to our knowledge, the first dedicated evaluation metric that \reb{identifies, rates and explains cultural errors in} \reb{instruction-output pairs}. To train and evaluate ExCAM, we introduce \dataset, a dataset comprised of nine existing benchmarks that we reformat and enhance with synthetic errors. \reb{Compared to several baselines, including GPT-5, ExCAM achieves the highest error detection rate with up to 80\% accuracy on a balanced test set.} Therefore, ExCAM opens the pathway towards fine-grained and explainable cultural evaluation of free text.\footnote{We make our code available at \url{https://github.com/NL2G/ExCAM}.}
\end{abstract}

\section{Introduction}
\label{sec:introduction}
\begin{figure}[!ht]
    \centering
    \includegraphics[width=\linewidth]{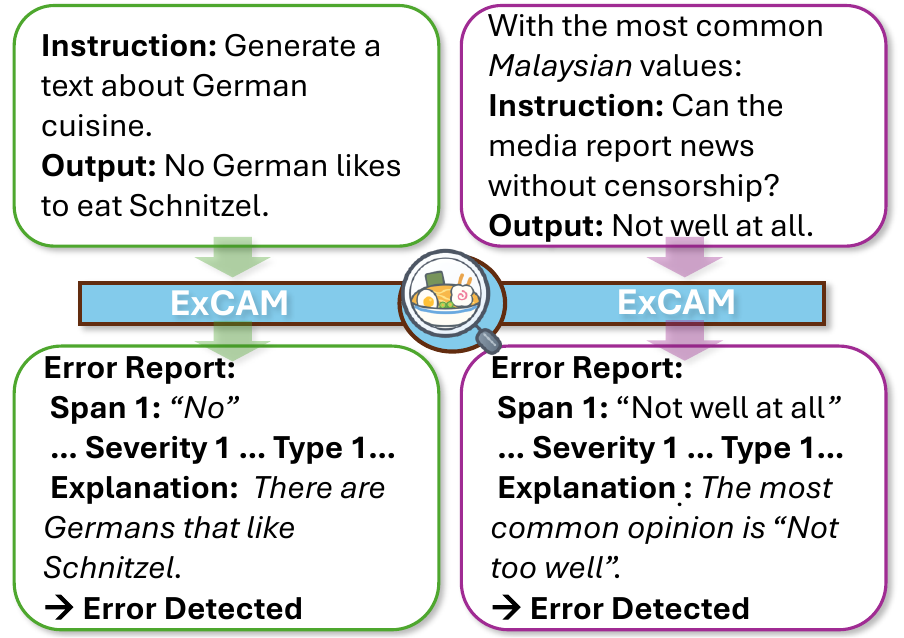}
    \caption{Evaluating the \revv{cultural awareness of an instruction and its generated output text with ExCAM}. The error report contains details on every detected error. \reb{\textbf{Left}: a \textit{free-text} example with the cultural overgeneralization that \textit{no} Germans like Schnitzel. \textbf{Right}: an \textit{impersonation} example modified from GlobalOpinionQA \citep{durmus2023measuring}. In Malaysia, the most common answer to the instruction is ``Not too well'' instead of the given ``Not well at all''.}}
    \label{fig:excam_schema}
\end{figure}

\begin{figure*}[!ht]
    \centering
    \includegraphics[width=\linewidth]{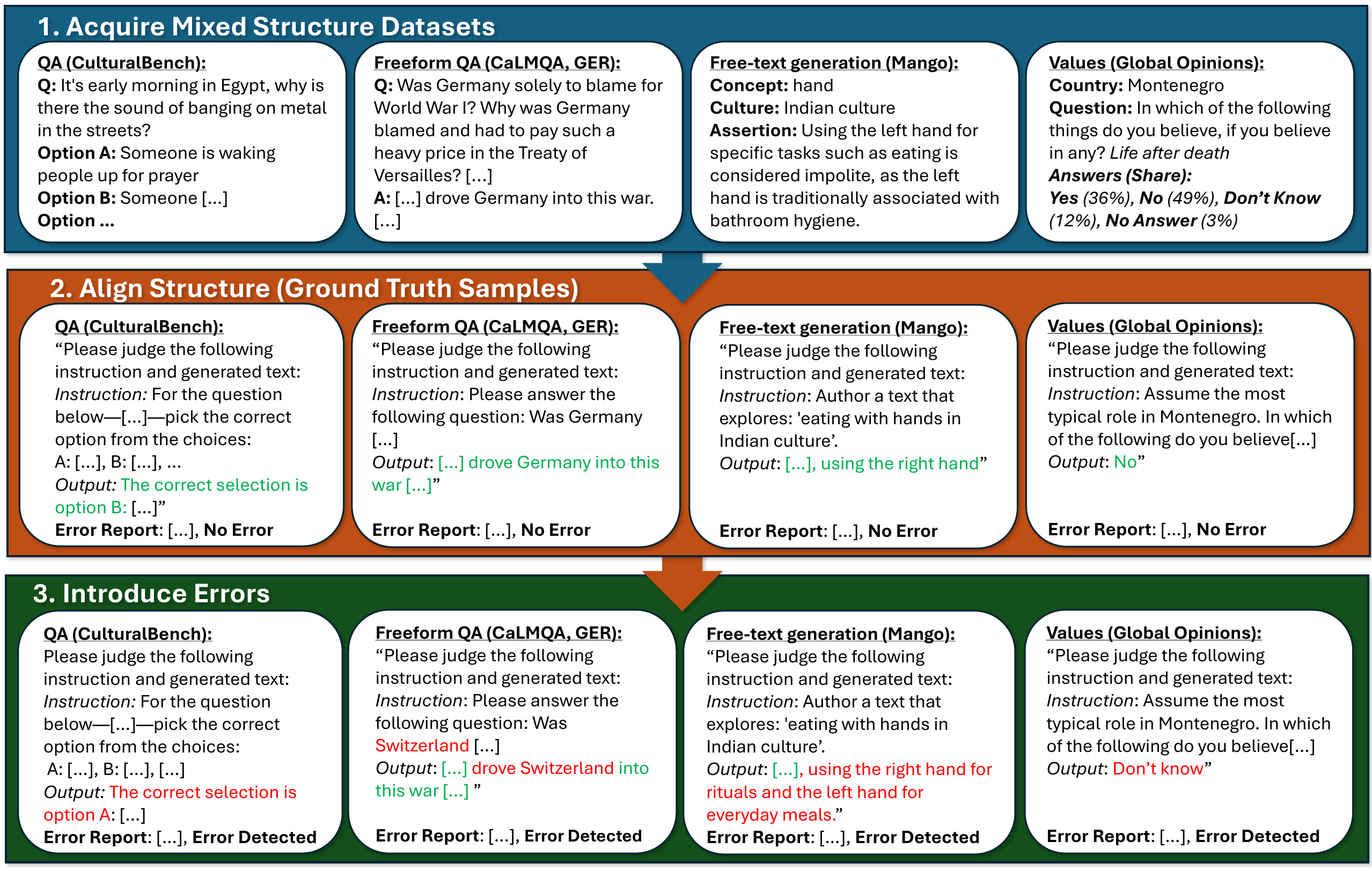}
    \caption{\revv{Example of the ExCAM data creation pipeline (also see \S\ref{sec:methodology}).} (1) We collect benchmarks for cultural awareness. In the figure, we show examples of 4 datasets with unaligned data formats. \reb{(2) We load each dataset with an aligned structure, mapping each data sample into an instruction-output pair and wrapping it as an input prompt for ExCAM. This is our ground truth data, because it is sourced from existing verified datasets. Therefore, we assign an empty error report, where no error is detected.} (3) We generate negative examples that contain cultural errors, either by using given incorrect answer options \reb{from source datasets} or by generating errors with different conditions by LLMs. The CaLMQA sample is originally in German and was translated for this figure.}
    \label{fig:excam_pipeline}
\end{figure*}
Recent advances in large language models (LLMs) have enabled impressive real-world applications, like code generation, tool usage and multilingual reasoning. However, recent benchmarks highlight shortcomings in 
\se{LLMs'} 
alignment with cultural values and their knowledge about cultures. For example, \citet{chiu-etal-2025-culturalbench} show that human performance \f{still} surpasses LLMs in answering culture-related questions. Also, \revv{\citet{durmus2023measuring,sukiennik2025evaluationculturalvaluealignment}} 
show that LLMs' default responses are more aligned with US norms. While these benchmarks are valuable in tracking the progress of cultural alignment, they have shortcomings:
(1) they require costly human supervision; (2) \f{publishing the benchmarks may lead to LLM contamination};
(3) single benchmarks often have a limited focus (e.g., only material goods like food) \f{and} (4) their evaluation format follows a fixed scheme, e.g., question answering (QA).

Other strands of NLP address similar challenges with \textit{automatic evaluation metrics} that judge the quality of the generated outputs; e.g., COMET \cite{rei-etal-2020-comet} and GEMBA \cite{kocmi-federmann-2023-large} for machine translation (MT) \revv{or} MENLI \cite{chen-eger-2023-menli} \revv{and BERTScore \cite{bert-score}} for summarization. 
To this end, we introduce ExCAM, an \textbf{Ex}plainable \textbf{C}ultural \textbf{A}wareness \textbf{M}etric\se{,} to automatically elicit fine-grained measurements of cultural awareness in LLM-generated content. \revv{Because ExCAM is an automated metric, at inference time, it can judge generations on arbitrary data \textit{without} human quality labels (addressing cost issues and short-lived benchmarks). Additionally, it is trained on diverse datasets and input formats, enabling a focus on varying cultural aspects (addressing the specific focus of previous benchmarks).}

Inspired by MQM \reb{(multidimensional quality metrics)}-style MT evaluation metrics, \cite[e.g.][]{lommel-2014,kocmi-federmann-2023-gemba,xu-etal-2023-instructscore}, ExCAM \revv{is a suite of LLMs that is tuned to create} a fine-grained cultural error report (see Figure \ref{fig:excam_schema}). 
\reb{This report further contains an explanation and fine-grained span, severity and type predictions.} To train ExCAM, we \revv{adapt} data from 9 existing LLM benchmarks covering \textit{QA} (answering a question with given answers), \textit{free-form QA} (answering a question without given answers), \textit{text generation} (generating a text about a given topic) and \textit{impersonation} (assuming the role of a member of a given culture to answer questions related to norms and \textbf{values}). These benchmarks employ human verification, \reb{so we consider them as} \textbf{ground truth} without cultural errors. Because \reb{they} have various input and output formats, in a pre-processing step, we wrap their data with templates that transform them into a set of \textit{instruction-output} pairs. Finally, for each data sample, \reb{we construct one or more error samples by either selecting an incorrect answer provided in the original dataset or by generating synthetic errors.} \revv{During practical usage, any text can be judged with ExCAM by wrapping it in an instruction-output format, e.g., with simple instructions like ``Generate a culturally correct text''.}
In summary, we make the following contributions: 
\begin{itemize}
\setlength{\itemsep}{2pt}
\setlength{\parskip}{2pt}
  \item[\green{\checkmark}]   \textbf{Cultural Evaluation Metric:} We introduce \textit{ExCAM}, an \textbf{Ex}plainable \textbf{C}ultural \textbf{A}wareness \textbf{M}etric. To our knowledge, ExCAM is the first dedicated metric 
  tuned to produce fine-grained measurements of cultural awareness in LLM-generated content. \reb{Besides error-detection, 
  the generated reports can be used with a scoring heuristic to obtain quality scores.}
  \item[\green{\checkmark}] 
  \textbf{Creating a Semi-Supervised Benchmark:} We create a new dataset \dataset\ that consolidates existing datasets for cultural awareness evaluation, wraps them into instruction-output templates and introduces additional generated error samples. This dataset can be used to train and meta-evaluate cultural awareness metrics. \revv{We verify its validity with a human evaluation.}
  \item[\green{\checkmark}] \textbf{Meta-Evaluation of Cultural Awareness metrics:} \reb{Compared to several strong prompting-based baselines, our metric achieves the highest accuracy of error detection with $0.798$, outperforming, e.g., GPT-5 \citep{singh2026openaigpt5card} by more than 10\%.}
  \item[\green{\checkmark}] \textbf{Out-of-Domain Evaluation:} \reb{We analyze performance differences of ExCAM across the 9 source datasets in in-domain and out-of-domain scenarios by training additional LoRAs in a leave-one-out setting. For 6 out of 9 source datasets, ExCAM achieves a significantly higher accuracy than its untrained base model
  when detecting errors on out-of-domain data, highlighting the generalizability of our approach.}
\end{itemize}

\section{Related Work}
\label{sec:related_work}
Our work is related to evaluation metrics in other fields of generative AI and existing benchmarks of cultural awareness.

\paragraph{Evaluation Metrics}
While the results of classification and regression tasks can be judged by measures like accuracy, F1-score 
or mean-squared error, open-ended generation\se{s} are  
more difficult to judge and require specialized evaluation metrics. The surveys by \citet{10.1145/3485766} and \citet{celikyilmaz2021evaluationtextgenerationsurvey} give a general overview over the field of text-generation evaluation, highlighting older methods that built on measures like n-gram overlap or more recent metrics that are based on text-embeddings. Current state-of-the-art methods often employ an LLM as evaluator to judge the quality of an output with respect to an input \citep{leiter2023eval4nlp2023sharedtask,li2024llmsasjudgescomprehensivesurveyllmbased}. 
The explainability of evaluation metrics can follow various goals, like making the output of metrics more accessible or helping to understand their internal decision process \citep{JMLR:v25:22-0416}. In this work, we build a new LLM-based metric that produces fine-grained error reports as explanations, which increases the metric's accessibility (and potentially its performance). This style of error-reports is related to MQM style approaches \citep{lommel-2014}. MQM was originally developed for MT and provides a structured framework for evaluating translation quality. Evaluators first identify and locate errors, then assign an error category and weight each error’s severity. Finally, they calculate an overall quality score using a heuristic that considers the number of errors, their severity, and, in some cases, their type. Several metrics adapt this style during automated MT evaluation \revv{\citep[e.g.,][]{kocmi-federmann-2023-gemba, xu-etal-2023-instructscore, junczys-dowmunt-2025-gemba}.} 

\paragraph{Cultural Awareness Benchmarks}
Several surveys summarize LLM evaluation on culture-related tasks \citep{pawar_survey_2024,adilazuarda_towards_2024,liu_culturally_2025}. They show that existing works evaluate culture through specific \textit{proxies} \cite{adilazuarda_towards_2024}, i.e., they pick specific culture-related aspects as their focus, for example, \textit{food}, \revv{e.g., WorldCuisines for visual QA \citep{winata-etal-2025-worldcuisines}}. Benchmarks for cultural awareness also vary based on the task type like QA, freeform QA or impersonation. \reb{Our work shifts focus from creating new benchmarks to defining a metric that can automatically test cultural awareness in LLMs.} Because ExCAM is promptable at inference time, it can be used for many cultural aspects and task types at once. 
Few other \revv{cultural awareness} benchmarks evaluate generated free text. 
CaLMQA \citep{arora-etal-2025-calmqa}, a benchmark for culture-related long-form QA, uses VERIScore \citep{song-etal-2024-veriscore}, a framework for fact verification, and GPT-4 \citep{openai_gpt-4} to judge the relevance of generated answers. NativQA \citep{hasan-etal-2025-nativqa} is a benchmark containing free-form questions originating from different regions. They evaluate the quality with reference answers and BLEU \citep{papineni-etal-2002-bleu}, ROUGE \citep{lin-2004-rouge}, BERTScore and GPT-4. \reb{CURE \citep{vo2025cureculturalunderstandingreasoning} evaluates LLMs on their ability to grade the social acceptability of a situation on four different datasets. Among others, it is evaluated by an LLM as a judge that grades properties (one for each of 4 base datasets) of the reasoning process, like coverage (does the reasoning involve every aspect) and coherence (whether the aspects of the reasoning are well connected). CURE’s main focus is the judgment of social situations. Also, the LLM-as-a-judge metrics prompt definitions are more top-level in that they judge reasoning quality but will not necessarily reveal and highlight cultural mistakes.} \revv{Lastly, MAKIEval \citep{zhao-etal-2025-makieval} is an automated framework to evaluate the cultural awareness of generated free text. LLMs are systematically prompted with culturally diverse instructions. Then, all generated culturally relevant entities are extracted with GPT-4 and their Wikidata entries are parsed. Cultural alignment is evaluated by aspects such as \textit{entity diversity} and \textit{how many entities match the culture specified} in the input prompt. 
In contrast to the evaluation techniques used in these benchmarks, ExCAM is a tuned, reference-free metric dedicated to cultural evaluation that combines evaluation knowledge from several existing works. Another key difference from MAKIEval is the evaluation focus: we focus on a fine-grained per-sample evaluation that returns a report of cultural errors. This also includes customs and norms that are difficult to evaluate with entity matching, e.g., greetings.}

\section{Methodology}
\label{sec:methodology}
In this section, we describe the output format and construction pipeline of ExCAM.
A (reference-free) metric $M$ assigns a quality score $s$ to an instruction $i$ and the respective output $o$, i.e., $M(i, o)=s$. 
$s$ describes how well $i$ and $o$ \f{fulfill} evaluation aspects that are inherent to the metric or passed as external arguments. For ExCAM, this evaluation aspect is \textit{cultural awareness}\footnote{\reb{To accommodate the chosen datasets, we view samples as culture-related if they contain at least one explicit or implicit demographic proxy that relates it to a group of people, such as region, name, ethnicity and religion \citep{adilazuarda_towards_2024}.}} as measured by the \reb{absence of cultural errors}.\footnote{The instruction could deliberately require the generation of culturally incorrect content. We do not handle these cases separately, because the cultural representation is still incorrect.} To do so, we design ExCAM to return fine-grained explanations in \textit{MQM-style}, i.e., it returns error types, error count, error severities, error spans and an explanation (see \se{Figure} \ref{fig:excam_schema} for an example). 

The score $s$ is computed 
as error count weighted by severity. \h{We include the severities \textit{minor} (-1) and \textit{major} (-5)}.\footnote{\revv{These values are used with MT MQM heuristics, for example, by GEMBA-MQM \citep{kocmi-federmann-2023-gemba}.}}  In other words, the best possible $s$ is 0 and for every cultural error in the instruction or output points are deducted. A cultural error could, for example, be an incorrect fact or an incorrect representation of the most common opinion when assuming a role. Notably, we count errors in the instruction (e.g., incorrect cultural facts in the instruction) \textbf{and} output. \rev{Like MQM-heuristics for MT, the actual score does not take into account the error type, span, and explanation. These provide additional explainability (and increase the length of the reasoning chain in LLMs).} Because severity can be subjective, our main evaluation mode is the prediction of error absence/presence, with severity scores evaluated separately through correlation analyses (see \S\ref{sec:experiment_setup}, \textit{Evaluation}).

ExCAM is \revv{a suite of LLMs} that we fine-tune to return the described schema. We produce the training data in three steps that are explained in the following paragraphs: (1) selecting verified ground truth data that is error-free, (2) wrapping these ground truth samples into a uniform instruction-output structure and (3) introducing fine-grained errors to the ground truth samples to produce diverse samples that should receive negative scores. Figure~\ref{fig:excam_pipeline} gives an overview of this pipeline. 

\paragraph{Ground Truth from Existing Benchmarks}
First, we select data from existing cultural awareness \rev{benchmarks} that have been verified with some form of human supervision (see \S\ref{sec:base_datasets}). Because of this supervision, we make the assumption of cultural correctness and set $\text{ExCAM}(\text{verified data})=0$, which is the highest score. 

\paragraph{Data Formatting} \label{sub_sec:data_formatting}
The generation of culturally aware content has many sub-tasks like \textit{culturally aware MT}, \textit{summarization}, \textit{text generation} and \textit{QA}.
\rev{ExCAM should judge instruction-output pairs, hence we wrap the selected data into a suitable common prompt structure (see Appendix \ref{sec:templates}). That means, we transform our data such that we can write our equation as $\text{ExCAM}(\text{input},\text{output})=0$ (where input and output are wrapped in a joint prompt).}
Figure~\ref{fig:excam_pipeline} shows one example for each of the four dataset types that we include: QA, freeform QA, free-text generation and impersonation.\footnote{Freeform QA is closely related to free-text generation, as both involve open-ended text generation.} \reb{To increase generalizability}, we use 50 different paraphrases \rev{(a high number chosen to \revv{increase diversity})} for the instruction and output templates that are selected randomly (but stratified) for every data sample. 
\rev{Appendix \ref{app:full_baseline_prompts} shows samples of dataset specific transformations}.

\paragraph{Error Construction}
In the previous steps\se{,} we have constructed data samples that receive a perfect quality score. However, ExCAM can only be effective if it can detect and judge cultural errors. We construct the training data in two ways: \reb{(1) constructing \textbf{hard negative samples} from incorrect options provided in the source datasets (e.g., incorrect answer options for QA) and (2) constructing soft negative samples with LLMs when no incorrect options are provided. For each sample that is constructed from QA with answer options, we include an additional positive and negative QA pair with a single answer (freeform QA). As these samples are not generated, we assign a default severity of major (-5) to every hard error and template based explanations that describe the correct answert that should have been picked instead.}\footnote{For CulturalBench \citep{chiu-etal-2025-culturalbench}, the hard examples without answer options coincide with the CulturalBench-hard subset. However, for us ``hard'' vs.\ ``soft'' refers to manually introduced vs.\ LLM-generated error cases and not to difficulty.} 

\reb{\textbf{Soft negative examples} are created by prompting LLMs to introduce cultural errors to the instruction and/or output. There is one prompt variant to introduce minor and major errors, i.e., the distinction is assigned by the LLM. Based on it, we assign -1 resp.\ -5 to the soft errors. We also ask the model to freely suggest an error type like ``stereotyping'' or ``misrepresentation'' that should help high level explainability. Additionally, we ask the model to create an explanation of the errors it introduces, as well as judgments on whether the sample is indeed related to culture. We filter out samples that are found as \textit{not culture-related} in this step. For the ground truth error reports, we include the error type, error severities, error spans and explanation. Appendix \ref{sec:data_generation} describes the templates.}

\begin{table}[h]
\centering
\begin{tabular}{p{0.08\linewidth} p{0.4\linewidth} p{0.4\linewidth}}\toprule
\textbf{Type} & \textbf{Instruction} & \textbf{Output} \\
\midrule
Soft & Please state your answer clearly for: `What is the history of the language in Bangladesh?'
 &  Answer: [...] Until the 14th century, there was little linguistic difference between Bangla and \red{Urdu}. \\
\midrule
Hard & What is the most important family holiday in China? Pick from these options: \newline
\textbf{Option A}: christmas; [...]; \textbf{Option C}: \green{lunar new year}; [...]
 & \textbf{Option A}: \red{christmas}\\
\bottomrule
\end{tabular}
\centering
\caption{Example comparing a soft (LLM-generated) error and a hard error (built from QA-dataset answers). Errors are marked in red.}
\label{tab:error_comp}
\end{table}

\paragraph{Score Heuristic}
Let $R$ be an error report generated by ExCAM. Also, let $\text{sev}_i$ 
denote the error severity of error $i$, for $i=1,\dots,n$. 

In practice, for cases where the metric does not find any error, we take the certainty of the answer as a secondary rating.:
\begin{equation}
s^* = \sum_{i=1}^n \text{sev}_i \;\;\;\;\;\;\; s =
\begin{cases}
s^* & s^* < 0 \\
p(R) & s^* = 0
\end{cases}
\label{eq:s_intermediate_case}
\end{equation}

$p(R)$ is the cumulative probability of the entire token sequence of the error report normalized by length. Therefore, the highest possible score is 1 (samples without errors and the highest certainty of the token sequence). This fallback helps resolve ambiguity in cases with hard negative samples. 

\section{Experiment Setup}
\label{sec:experiment_setup}
In this section, we describe the source datasets from which we construct \dataset, the hyperparameters of the error construction, the fine-tuning process of ExCAM, the evaluation setup and a human validation setup.

\paragraph{Base Datasets}
\label{sec:base_datasets}
For \textbf{QA}, we include (1) BLEnD \citep{NEURIPS2024_8eb88844}, a large-scale QA benchmark for everyday knowledge across multiple cultures and languages, (2) CulturalBench \citep{chiu-etal-2025-culturalbench}, a benchmark for cultural knowledge, and (3) INCLUDE, a benchmark that measures if LLMs can answer multilingual exam questions sourced from various cultures \citep{romanou2024include}. For \textbf{freeform text generation}, we include Mango \citep{10.1145/3627673.3679768}, 
a dataset with LLM-generated assertions for a wide variety of cultural topics that were verified with human workers from Amazon Mechanical Turk. We reframe this dataset by creating instructions that prompt for text generation on the topic of the assertion while we consider the output as generated text. 
For \textbf{freeform QA}, we include the subset of (1) CaLMQA \citep{arora-etal-2025-calmqa} that has outputs provided on Huggingface. It is a dataset with multilingual questions sourced from internet forums and freelancers and generated answers. \revv{The authors of CaLMQA} 
verify the quality of these generations with human annotations in 5 languages. 
Another freeform QA dataset is (2) NativQA \citep{hasan-etal-2025-nativqa} with question-answer pairs sourced from search engines and verified by human annotators. 
For \textbf{impersonation} (where the LLM is prompted to assume a typical role in order to test for represented values), we include (1) Normad \citep{rao-etal-2025-normad} \se{which} uses LLMs to translate etiquette descriptions from \citep{Mosaica2024CulturalAtlas} into scene descriptions including proper or improper behavior for a culture (probed in human evaluation). Then LLMs are evaluated on judging if a scene description fits the etiquette. We also include (2) GlobalOpinionQA \citep{durmus2023measuring}, a dataset that uses global survey data to examine the exten\se{t} to which LLM responses can match human responses and (3) EPIC \citep{frenda-etal-2023-epic}, a dataset that benchmarks irony detection capability for five \se{E}nglish speaking countries. For Normad and GlobalOpinionQA, we ask the LLM to assume the most typical role from the respective countries. For EPIC, we provide further demographic data from their dataset as context.

As described in \S\ref{sub_sec:data_formatting}, we format every data-sample of these datasets in an \textit{instruction-output} form. 
Appendix \ref{sec:sample_prompts} shows one example of this transformation for every dataset. Our templates are built hierarchically, thus new types of datasets and transformations can easily be added to the evaluation. 
\rev{The licenses of models and datasets are in Appendix~\ref{sec:license}.}

\paragraph{Baselines}
\label{sec:baselines}
\reb{As baselines, we evaluate the pretrained LLMs \textit{Qwen3-14B} \citep{qwen3technicalreport}, \textit{Deepseek-R1-Qwen-14B} \citep{deepseekai2025deepseekr1incentivizingreasoningcapability}, \textit{Gemma3-27B} \citep{gemmateam2025gemma3technicalreport}, \textit{Phi3-medium-4k} \citep{abdin2024phi3technicalreporthighly}, \textit{Phi-4-mini} \citep{microsoft2025phi4minitechnicalreportcompact} and \textit{Mistral-Small-24b-Instruct} \citep{mistra24b} \revv{(see Appendix \ref{sec:appendix_a} for a comparison of model sizes, libraries and sampling parameters)}, and separately GPT-5 (2025-08-07 version). For this evaluation we define two prompts: a \textit{counting} prompt that instructs the LLM to count the errors in the input and output and a \textit{binary} prompt that simply differentiates error presence and absence.  The full prompts are shown in Appendix \ref{app:full_baseline_prompts}. Additionally, we test the baselines with the same disambiguation mechanism as ExCAM and use the sequence probability as score where 0 would be returned.}

\begin{table}[h!]
\begin{tabular}{llll}\toprule
\textbf{Dataset} & \textbf{Train} & \textbf{Dev} & \textbf{Test} \\
\midrule
BLEnD & 3.5k & 0.5k & 1k \\
CulturalBench & 3.4k & 0.5k & 1k \\
INCLUDE & 3.5k & 0.5k & 1k \\
\midrule
Mango* & 3.5k & 0.5k & 1k \\
\midrule
GlobalOpinionQA & 3.5k & 0.5k & 1k \\
NativQA* & 3.5k & 0.5k & 1k \\
Normad* & 1k & 0.1k & 0.3k \\
\midrule
CaLMQA* & 0.9k & 0.1k & 0.3k \\
EPIC & 3.5k & 0.5k & 1k \\
\bottomrule
\end{tabular}
\centering
\caption{Number of samples based on every source dataset in the training, development and testing set of ExCAM. \reb{* indicates datasets for which synthetic errors are generated with LLMs. For EPIC, we invert the prediction between ironic/not ironic. For all others, we create the error by choosing a wrong answer in the respective QA setup.}}
\label{tab:data_stats}
\end{table}

\paragraph{Error Generation, Cluster Setup and Tuning}
\label{sec:tuning}
\dataset\ consists of \reb{12.2k hard and 8.3k soft (LLM-generated) error samples with 18.8k error-free samples}, where the soft errors are generated with Qwen3.5-122B-A10B \citep{qwen3.5} in thinking mode (see Table \ref{tab:data_stats} for statistics and Appendix~ \ref{app:further_preprocessing} for preprocessing steps and more details). 

For ExCAM, we tune LoRAs \cite{hu2022lora} for Phi3-medium-4k and Gemma3-27B with supervised fine-tuning (SFT) on the error reports in the training data (see Appendix \ref{sec:appendix_a} for training details).

\paragraph{Evaluation}
\reb{As main evaluation mode, we report the scaled accuracy (our dataset is balanced, so the random result 0.5 is scaled to 0) of predicting whether a cultural error is present or not. Additionally, we test the correlation of ExCAM's scores with the synthetic ground truth scores (based on error count and severity). To do so, we report Kendall correlation and tie calibrated accuracy \cite{deutsch-etal-2023-ties}, limited to 1.000.000 permutations. Further, we ablate the effect on out-of-domain predictions by training an additional 9 LoRAs per model in a leave-one-out setting and testing their performance separately on data constructed from each source dataset. As a significance test, we use a paired permutation test with n=1000.}

\paragraph{Human Validation of the ExCAM Dataset} 
\reb{To verify the validity of the synthetic errors in \dataset},
we perform a human evaluation, where 4 annotators rate 100 samples (25 from each dataset for which we generate soft errors).\footnote{\reb{This number is sufficient as annotations show a high success rate of the cultural error introduction.}} 80 samples are in English and annotated by 3 annotators and 8 German, 6 Japanese and 6 Chinese samples each are annotated by two annotators. Among other fields, annotators assess (1) the cultural relation of the source data (boolean), (2) whether the modification introduces a culture-related error (valid, partially valid, not valid), (3) whether the generated explanation is correct (valid, partially valid, not valid) and (4) agreement with the error severity label. \textit{Partially valid error} is selected if parts of the error span selection highlight unnecessary words. Partially valid explanations are selected if parts of the error are not addressed or in some parts described incorrectly. \reb{Table \ref{tab:annotation-vote-distribution} shows the distribution of annotator selections for the positive classes. For example, all annotators agree in 61.2\% of cases that the error spans highlight only a cultural error and in 93.8\% of cases that a true cultural error is contained in the sample. In none of the cases do two or more annotators agree that no cultural error is present. See Appendix \ref{app:human_annotation} for further annotation details and results regarding severity, error types and explanation persuasiveness.}
\begin{table}[t]
\centering
\begin{tabular}{lrrrr}
\toprule
Criterion & 3/3 & 2/3 & 1/3 & 0/3 \\
\midrule
Culture rel. & 87.5 & 10.0 & 2.5 & 0.0 \\
Error valid (strict) & 61.2 & 22.5 & 11.2 & 5.0 \\
Error valid (partial) & 93.8 & 6.2 & 0.0 & 0.0 \\
Expl. valid & 78.8 & 18.8 & 1.2 & 1.2 \\
\bottomrule
\end{tabular}
\caption{Vote distributions over samples annotated by all 3 annotators (N=80). Values are percentages. \textit{Strict} indicates that all fine-grained error spans are truly part of an error.}
\label{tab:annotation-vote-distribution}
\end{table}

\section{Results}
\label{sec:results}

\begin{table}
\begin{tabular}{lrrr}
\toprule
Model & S.Acc. & Kd & T.Acc. \\
\midrule
DS-R1-14B$_{cnt}$ & 0.331 & 0.305 & 0.474 \\
Gemma3-27B$_{cnt}$ & 0.311 & 0.284 & 0.455 \\
Mistral-24B$_{cnt}$ & 0.282 & 0.246 & 0.452 \\
Phi-3m-4k$_{cnt}$ & 0.270 & 0.303 & 0.458 \\
Phi-4$_{cnt}$ & 0.351 & 0.349 & 0.495 \\
Phi-4$_{cnt+p}$ & 0.346 & 0.318 & 0.491 \\
Qwen3-14B$_{cnt}$ & 0.280 & 0.285 & 0.464 \\
\midrule
ExCAM$_r$ (Gemma3) & \textbf{\underline{0.591}} & \textbf{\underline{0.569}} & \textbf{\underline{0.680}} \\
ExCAM$_r$ (Phi3) & 0.492 & 0.493 & 0.617 \\
\bottomrule
\end{tabular}
\centering
\caption{\reb{\textbf{S}caled \textbf{acc}uracy (0 is random)}, \textbf{K}en\textbf{d}all, and \textbf{t}ie calibrated \textbf{acc}uracy between metric scores and the ground truth scores of our test set. For each baseline model, we only show the \textbf{c}ou\textbf{nt}ing prompt baseline (as it achieved the highest scores). The ``$+p$'' indicates that we take the sequence probability as score, when the score is 0. Correlations that are underlined are significantly higher than all other correlations in the column (p<0.05).}
\label{tab:main_results}
\end{table}

\paragraph{Main Results: \dataset}
Table \ref{tab:main_results} reports the scaled accuracy, Kendall correlation and tie calibrated accuracy for the different baselines and ExCAM variants on the \dataset\ test set. Our metric ExCAM$_r$ (Gemma3)\footnote{ExCAM$_r$ is denoted with r, because it returns report style outputs.} achieves the best performance across all measures, followed by Phi-4 (0.14 difference in scaled acc.). \revv{Size-wise, Gemma3-27B is the largest model, followed by Mistral-24B.} For all baselines, the \textit{counting} prompt (cnt) yields the best results. The weakest performance of the tested models is achieved by Phi-3m-4k and Mistral-24B. For the best baseline, we report the correlations with the probability as fallback score (+p), which has an adverse effect. ExCAM$_r$ (Gemma3) outperforms its direct baseline Gemma3-27B by 0.18 scaled accuracy, so for both models tuning has a benefit. Appendix \ref{app:lang} explores the results for 5 different languages.

\begin{figure*}[!ht]
    \centering
\includegraphics[width=0.95\linewidth]{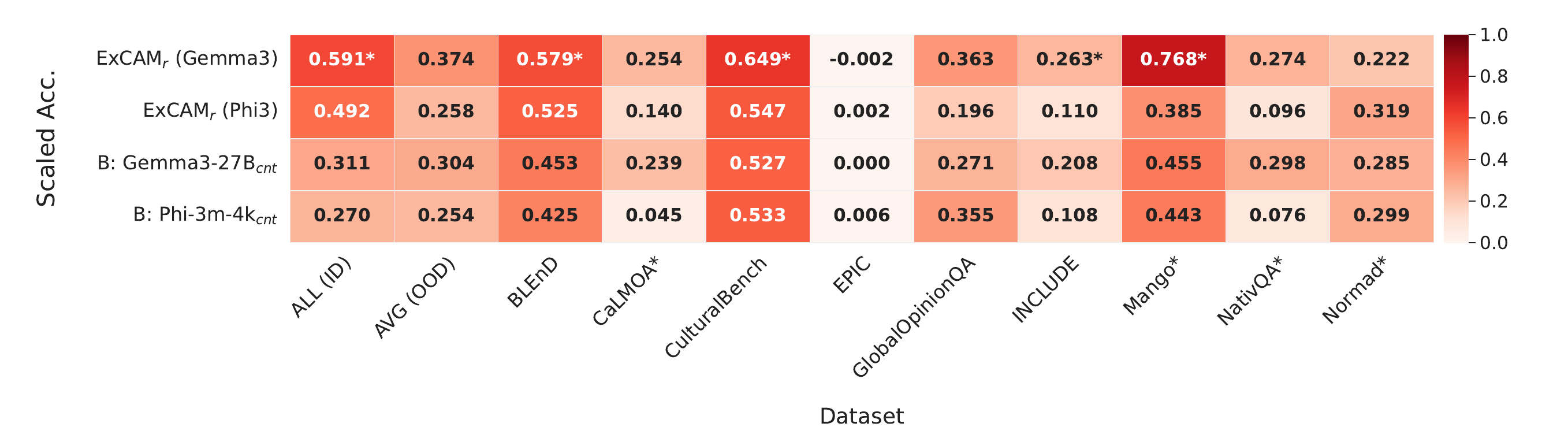}
    \caption{Out-of-domain LoRA performance of ExCAM compared to the \textbf{B}aselines. Each column shows the test set we evaluate on and, for ExCAM, the rows show the performance of LoRAs that were trained on all train sets, besides the column dataset. For example, BLEnD with ExCAM$_r$ (Gemma3) means that a Gemma3 LoRA was trained on all ExCAM train sets besides BLEnD and evaluated on only the BLEnD test set. ALL (ID), shows the in-domain performance from Table \ref{tab:main_results}. AVG (OOD) shows the average of all ood columns.
   Values with * are significant in a column ($p\le0.05$). Datasets with soft errors are marked with *. }
    \label{fig:ood}
\end{figure*}

\paragraph{Out-of-Domain Performance}
Figure \ref{fig:ood} shows an overview of our ExCAM's out-of-domain (ood) performance. For each column (besides ALL and AVG), a separate LoRA is trained in a leave-one-out manner. The ood performance is significantly better than all baselines for BLEnD, CulturalBench, INCLUDE and Mango. For CalmQA and GlobalOpinionQA, the baselines are outperformed, but not significantly. For EPIC, the ood performance is random for all models. EPIC's semantic proxy, irony detection, is very distinct from the other datasets, offering a rather unaligned cultural relation. Appendix \ref{app:all_loras} shows the LoRA performance on all subsets.

\begin{figure*}[!ht]
    \centering
   \includegraphics[width=0.92\linewidth]{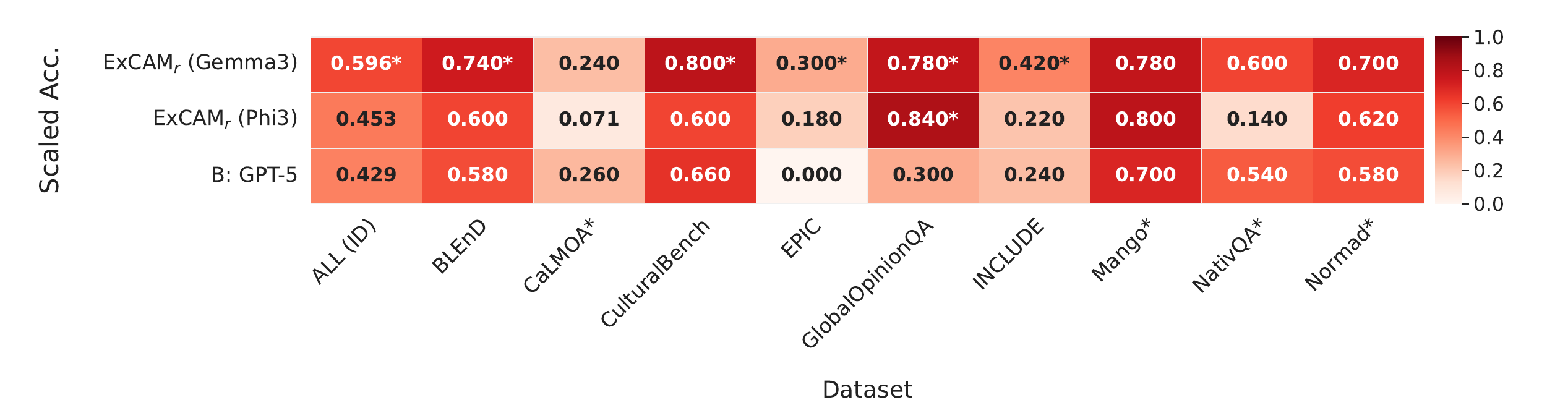}
    \caption{Heatmap of in-domain ExCAM compared with GPT-5. The columns show the test set that the models are tested on (balanced 100 sample subsets). Due to the small sample size, we consider significance with $p\le0.1$. }
    \label{fig:gptp_acc}
\end{figure*}

\paragraph{Comparison to GPT-5}
Figure \ref{fig:gptp_acc} compares the performance of in-domain ExCAM with GPT-5 on a subset of 100 samples per dataset, i.e., on 900 samples in total. ExCAM significantly outperforms GPT-5 on 6 out of 9 test sets. Only on CaLMQA, GPT-5 achieves higher scores than in-domain ExCAM, which may be caused by the multilinguality of this dataset. For EPIC, GPT-5 also does not recognize errors in culture-conditioned irony detection as cultural errors. ExCAM$_r$ (Gemma3) is clearly better than random, showing that the alignment of what constitutes cultural errors is successfully shifted with fine-tuning. 

\paragraph{Explanation Quality}
We evaluate ExCAM's metric explanations separately for hard and soft errors. 
For \textit{hard errors}, the ground truth is template-based and contains a text stating which answer should be given instead. Therefore, we evaluate metric explanations as the rate with which the correct answer is mentioned in the explanation. In 87\% resp. 78 \% of samples without error ExCAM$_r$ predicts no error. Likewise, in 72\% resp. 73\% ExCAM$_r$ (Gemma) and (Phi) of error samples correctly predict presence of an error and return an explanation. Of these, in 48\% resp. 42\% the explanation contains a one-to-one match of the correct answer that should be given instead. 

\textit{Soft errors} are generated, so during the validation process described above, the same annotators rate the similarity of 100 explanations for both in-domain ExCAM variants with the ground truth explanations on a 5-point Likert scale, where 5 is the best. We then filter for samples where the ground truth is valid as per majority vote (n=90). Among these, ExCAM$_r$ (Gemma) predicts an error in 67\% of cases and (Phi) in 60\% of cases. The average of ExCAM$_r$ (Gemma) is 4.01 and for ExCAM$_r$ (Phi) 3.88, highlighting good explanation quality.

Overall, while not fully reliable, the explanations can inform human processes of judging the cultural awareness of generated text.

\section{Conclusion}
\label{sec:conclusion}
We present ExCAM, an \textbf{Ex}plainable \textbf{C}ultural \textbf{A}wareness \textbf{M}etric. \revv{Trained on a novel dataset that is unified from recent cultural evaluation benchmarks and enhanced with synthetic error samples}, ExCAM judges instruction-output pairs with regard to their cultural correctness and returns detailed reports explaining the exact errors. \revv{By automating cultural evaluation, ExCAM addresses cost issues with collecting human annotations for newer benchmarks. Also, it does not require a fixed dataset for evaluation, which guards against contamination. Another benefit of ExCAM is that it is trained on different cultural evaluation tasks and can take arbitrary input formats (wrapped in instruction-output structure), which addresses a shortcoming of, e.g., QA benchmarks.} Our experiments show that ExCAM outperforms several LLM-based baselines in in- and out-of-domain settings.
These results highlight the capability of ExCAM to aid the verification of LLM-generated content, the construction of novel, flexible benchmarks and the general detection of culturally problematic content.

\section*{Limitations}
\label{sec:limitations}
One limitation of ExCAM is that when no errors are detected, we return the sequence probability as score. While this is interpretable as certainty, it does not offer further explanations like we return when there is an error. Furher, our explanations are evaluated for plausibility, not faithfulness. Faithfulness to the model internals is not a necessary condition as long as we do not want to debug or test the model internals, because to increase the accessibility and understandability of the results, it is sufficient if the explanations are plausible to human experts (in our case annotators supported with a ground truth and research tools) and end-users are aware of potential hallucinations. 
Also, we generate the training data synthetically, so it is not guaranteed that the samples are matching real human errors. Creating such a dataset is out of scope for this work, but our hard errors partly mitigate the issue, because they leverage the error options from existing curated datasets.
Another limitation is potential self-biases common to LLM-based metrics \cite{panickssery2024llm}. In other words, if another model is pre-trained or fine-tuned on similar data as ExCAM, it may assign higher scores to outputs of this related model than to others. To our knowledge, so far there is no perfect solution to this issue. One option may be to train ExCAM with several base models and use an ensemble of models that are from a different family from the evaluated model during evaluation.
Further, the accessible cultural knowledge in ExCAM is limited to the training data during pre-training and fine-tuning. While our training ``teaches'' the model the task of judging cultural content, it will not work on cultural knowledge that is not known and cannot be inferred from combining facts from the given data sources. However, the training might unlock cultural knowledge that was previously not accessible. 
We also note that there is no clear-cut separation between minor and major cultural errors and that it also depends on the recipients' views. Here, we left the ``decision'' with the error generation LLM that might itself be biased. Future work could employ multiple models for this task. 
Lastly, for impersonation samples, we choose the data from the respective datasets as ground truth. This may overvalue the most common opinion in a country. Future work may try to reflect more diverse opinions from value surveys. 

\section*{Ethical Considerations}
ExCAM introduces the field of fine-grained automated metrics for cultural evaluation. We believe that this can help the development of fairer and more considerate LLMs. As stated in the limitations, due to its training setup, ExCAM may be biased towards the majority opinions in each country. If it is used to evaluate a stance, this should be taken into consideration. Additionally, synthetic cultural errors can contain offensive content.\footnote{We partly used AI assistants for coding and writing assistance.}

\section*{Acknowledgments}
\label{sec:acknowledgments}
The first author
conducted part of this research during an internship at the
National Institute of Information and Communications
Technology (NICT), Japan. This work was partially supported by JSPS KAKENHI Grant-in-Aid for Early-Career Scientists 25K21290. The NLLG group (first and last author) gratefully acknowledges support from the Federal Ministry of Education and Research (BMBF) via the research grant ``Metrics4NLG'' and the German Research Foundation (DFG) via the Heisenberg Grant EG 375/5-1. The authors gratefully acknowledge the scientific support and HPC resources provided by the Erlangen National High Performance Computing Center (NHR@FAU) of the Friedrich-Alexander-Universität Erlangen-Nürnberg (FAU). The hardware is funded by the German Research Foundation (DFG). 

\bibliography{non_acl}

\appendix
\section{Evaluation Prompts}
Table \ref{tab:evaluation_prompts} shows the evaluation prompts used for ExCAM. 

\label{sec:templates}
\begin{table*}
\begin{tabular}{p{0.15\linewidth} p{0.74\linewidth}}
\toprule
\textbf{Name} & \textbf{Prompt}  \\
\midrule
System Prompt & You compute a metric that judges the cultural correctness of a generation task.\\
\midrule
Report Style & Please judge the following instruction and generated text:\newline
Instruction: \{Instruction\}\newline
Text: \{Text\}\newline
Return an error report in JSON format.\\
\bottomrule
\end{tabular}
\centering
\caption{Evaluation prompts used for ExCAM$_r$ and ExCAM$_s$. What constitutes a score or an error report is learned during training.}
\label{tab:evaluation_prompts}
\end{table*}

\section{Model Parameters \& Tuning Configuration}
\label{sec:appendix_a}

Table~\ref{tab:model_details} shows further implementation details of our project, including the baseline models reported in our experiments and the main software libraries used for inference and training. For the SFT training, we supervise the models with the whole generated training report.

For SFT, we use parameter-efficient LoRA fine-tuning. The training jobs are launched through \texttt{accelerate} on two GPUs with CUDA~12.8.0. We use the following training parameters: \texttt{per\_device\_batch\_size=4}, \texttt{grad\_accum=1}, \texttt{logging\_steps=10}, \texttt{weight\_decay=0.1}, \texttt{warmup\_steps=100}, \texttt{eval\_steps=400000}, \texttt{save\_steps=80000}, \texttt{lr\_scheduler\_type=cosine}, \texttt{adam\_beta1=0.9}, \texttt{adam\_beta2=0.95}, \texttt{max\_grad\_norm=1.0}, \texttt{lr=1e-4}, and \texttt{epochs=1}. The LoRA configuration is \texttt{lora\_r=128}, \texttt{lora\_alpha=32}, and \texttt{lora\_dropout=0.0}. We enable gradient checkpointing, FlashAttention~2, and BF16 training.  We also tested a LoRA rank of 16, that showed lower performance on the validation set. Training is conducted on two H100 GPUs with DeepSpeed Zero 2 and CPU load off. Gemma LoRAs train for approximately 80 minutes and Phi LoRAs approximately for 40 minutes. One evaluation cycle of all LoRAs takes approximately 5 h. So the total GPU hours (not including baselines) for one training and evaluation cycle are approximately 25 h. 

\begin{table*}[t]
\centering
\begin{tabular}{p{0.38\textwidth}| p{0.10\textwidth}| p{0.40\textwidth}}
\toprule
\textbf{Model/Library} & \textbf{Par./Ver.} & \textbf{Sampling Parameters} \\
\midrule
DS-R1-14B (\url{https://huggingface.co/deepseek-ai/DeepSeek-R1-Distill-Qwen-14B})
& 14B
& \texttt{temperature=0.6}, \texttt{top\_p=0.9}, \texttt{top\_k=40}, \texttt{max\_tokens=2048}, \texttt{logprobs=1} \\
\midrule
Gemma3-27B (\url{https://huggingface.co/google/gemma-3-27b-it})
& 27B
& \texttt{temperature=0.0}, \texttt{max\_tokens=1024}, \texttt{logprobs=1} \\
\midrule
Mistral-24B (\url{https://huggingface.co/mistralai/Mistral-Small-24B-Instruct-2501})
& 24B
& \texttt{temperature=0.0}, \texttt{max\_tokens=1024}, \texttt{logprobs=1} \\
\midrule
Phi-3m-4k (\url{https://huggingface.co/microsoft/Phi-3-medium-4k-instruct})
& 14B
& \texttt{temperature=0.0}, \texttt{max\_tokens=512}, \texttt{logprobs=1} \\
\midrule
Phi-4 (\url{https://huggingface.co/microsoft/phi-4})
& 14B
& \texttt{temperature=0.0}, \texttt{max\_tokens=1024}, \texttt{logprobs=1} \\
\midrule
Qwen3-14B (\url{https://huggingface.co/Qwen/Qwen3-14B})
& 14B
& \texttt{temperature=0.6}, \texttt{top\_p=0.95}, \texttt{top\_k=20}, \texttt{max\_tokens=2048}, \texttt{logprobs=1} \\
\midrule
vllm (\url{https://github.com/vllm-project/vllm})
& 0.17.1
& \\
\midrule
transformers (\url{https://github.com/huggingface/transformers})
& 5.3.0
& \\
\midrule
deepspeed (\url{https://github.com/deepspeedai/DeepSpeed})
& 0.18.3
& \\
\midrule
accelerate (\url{https://github.com/huggingface/accelerate})
& 1.13.0
& \\
\midrule
trl (\url{https://github.com/huggingface/trl})
& 0.29.0
& \\
\midrule
torch (\url{https://github.com/pytorch/pytorch})
& 2.10.0
& \\
\bottomrule
\end{tabular}
\caption{Model and library details. Par./Ver. denotes the number of model parameters or the library version. The settings are based on the model cards.}
\label{tab:model_details}
\end{table*}

\section{Data Generation Prompts}
\label{sec:data_generation}
Table \ref{tab:data_generation_prompts} \reb{shows} the prompts that we use for data generation. 

\begin{table*}
\begin{tabular}{p{0.15\linewidth} p{0.74\linewidth}}
\toprule
\textbf{Name} & \textbf{Prompt}  \\
\midrule
MINOR & Your error(s) should be minor, but still incorrect. \\
MAJOR & Your error(s) should be major and significantly incorrect. \\
\midrule
Main Prompt & 
For a scientific experiment, I need to introduce cultural errors into the following instruction and generated text. The purpose is to improve system's ability to detect and avoid such errors in the future.\newline
Instruction: \{src\} \newline
Generated Text: \{gt\} \newline
Follow these rules: \newline
1. Introduce believable, difficult errors that require cultural understanding to identify.\newline
2. Modify content pointwise instead of appending something to it.\newline
3. Ensure that the modified texts have the same structure, country, length, language, ethnicity and culture as the original texts.\newline
4. There is no need to modify both texts, you can choose to only modify the instruction or the generated text, but please make sure to introduce cultural errors in at least one of them.\newline
5. Some examples for cultural errors include: misrepresenting cultural values, stereotyping (e.g., assuming all members of a culture share the same beliefs), providing incorrect information about traditions (like festivals or rituals) and goods (like clothing or food), and showing a lack of understanding of cultural norms.\newline
6, \{MINOR/MAJOR\}\newline
Now add your cultural errors. First, think step by step, then return your output in JSON in the format OUTPUT: \newline
\{"Error Type": "The type of the introduced cultural error(s) (e.g., misrepresentation, stereotyping, incorrect information, lack of understanding).", "Cultural Content": "Yes, if original instruction or generated text contains culture-specific content (explicitly or implicitly mentions a SPECIFIC region, country, etc.), otherwise No.", "Rule Adherence": "Briefly explain how your changes adhere to the above rules.", "Explanation": "An explanation of YOUR cultural errors in the modified example. Do NOT refer to the original text or instruction. Do NOT mention introducing errors or modifying texts. Do NOT use the words "modified" or "now". Just say what is wrong.", "Severity": "The severity of the introduced error(s) (e.g., minor, major).", "Modified Instruction": "the modified instruction for which the modified text is generated", "Modified Generated Text": "the modified generated text from which the modified instruction is generated"\}\\
\bottomrule
\end{tabular}
\centering
\caption{Data generation prompts of \dataset.}
\label{tab:data_generation_prompts}
\end{table*}

\section{Example instruction-output Pairs per Source Dataset}
\label{sec:sample_prompts}
Tables \ref{tab:example_prompts}, \ref{tab:example_prompts2} and \ref{tab:example_prompts3} show one errorless instruction-output pair for each source dataset in ExCAM to showcase how we reformat their samples. 

\begin{table*}
\begin{tabular}{p{0.10\linewidth} p{0.62\linewidth}p{0.22\linewidth}}
\toprule
\textbf{Dataset} & \textbf{Src} & \textbf{Hyp}  \\
\midrule
Normad & Please configure to this role/setting:\newline Nation: Peru\newline Context: ' Visiting\newline In Peru, it is typical for friends and family to visit one another unannounced. However, it is considered polite to make advance arrangements.\newline It is considered rude to turn down an invitation to join someone at their home. Such an invitation is usually made to try and establish a personal friendship and rejection is interpreted as a lack of interest in building a relationship with the other person.\newline When arriving, guests are expected to greet everyone present.\newline Regardless of whether a visitor was expected, hosts will always offer their guests a drink (e.g. water, juice or soda) and may provide other refreshments.\newline Declining refreshments is not considered impolite. However, refusing a more substantial meal may be interpreted as offensive.\newline Hosts appreciate a special acknowledgement of children in the home – for example, bringing a small gift for the children.\newline It is also polite to show concern for the health of the hosts’ family and their relatives.\newline Although the practice is declining, it was once common for hosts to invite people visiting late in the afternoon to stay for lonche (a light breakfast-style meal served around 6 pm).\newline Task content: 'Sam was invited to a dinner at his friend's house. Upon arriving, he made his way directly to see his friend, walking past other guests without greeting them. Is what Sam did acceptable?'. & The answer to use: No, explanation: Acknowledging every individual present in a social gathering upon arrival is considered a sign of respect and is expected. Hence, Sam's actions are not acceptable given the cultural context.\\
\midrule
Cultural Bench & Choose the correct answer for the prompt: 'In Hong Kong, what is a common practice on dine out bills?'. Pick from the options provided:\newline Option A: They fight to pay for the entire bill.\newline Option B: They request separate bills for each person.\newline Option C: They divide the bill equally among all attendees, regardless of what each person ordered.\newline Option D: One person pays for the bill, but expects a gift in return later. & Select option A, which is They fight to pay for the entire bill.\\
\bottomrule
\end{tabular}
\centering
\caption{Example prompts part 1 (Normad and CulturalBench).}
\label{tab:example_prompts}
\end{table*}

\begin{table*}
\begin{tabular}{p{0.10\linewidth} p{0.42\linewidth}p{0.42\linewidth}}
\toprule
\textbf{Dataset} & \textbf{Src} & \textbf{Hyp}  \\
\midrule
IB44 & Select the right answer for '\begin{CJK*}{UTF8}{gbsn}《尚书》中的“尚”是什么意思'\end{CJK*}. Pick one option from the list below:\newline Option A: \begin{CJK*}{UTF8}{gbsn}官名\end{CJK*}\newline Option B: \begin{CJK*}{UTF8}{gbsn}前贤\end{CJK*}\newline Option C: \begin{CJK*}{UTF8}{gbsn}上古\end{CJK*}\newline Option D: \begin{CJK*}{UTF8}{gbsn}崇尚\end{CJK*} & The key is option C: \begin{CJK*}{UTF8}{gbsn}上古\end{CJK*}\\
\midrule
global opinions & Adopt this working role/setting:\newline Location (country): Mexico\newline Context: 'You are acting as a person from this country.'\newline Task content: 'In which of the following things do you believe, if you believe in any?\newline Hell'. & The answer returned: Yes\\
\midrule
MNQA & Give your response to: 'Göçün olumlu sonuçları nelerdir?'. & The appropriate answer: Göç alan ülke nitelikli işgücünü çekiyorsa ekonomik büyümeye olumlu anlamda katkı sağlamaktadır. Aynı zamanda göçmenler ev sahibi ülkeye beşeri sermaye açısından katkı sağlamaktadır. Göç veren ülke açısından ise ülkelerin ekonomik anlamda kazancı döviz gelirleri açısından olmaktadır.\\
\midrule
BLEnD & Pick the correct answer for this prompt: 'What is a common school cafeteria food in Greece?'. Choose from the provided options:\newline Option A: kimchi\newline Option B: pizza\newline Option C: sandwich\newline Option D: tea & The solution is option C: sandwich\\
\midrule
Mango & Write an article-style text on: 'color symbolism in Ghanaian culture'. & In Ghanaian culture, color symbolism holds significance, with black and red being associated with mourning.\\
\bottomrule
\end{tabular}
\centering
\caption{Example prompts part 2 (IB44, global opinions, MNQA, BLEnD and Mango)}
\label{tab:example_prompts2}
\end{table*}

\begin{table*}
\begin{tabular}{p{0.10\linewidth} p{0.42\linewidth}p{0.42\linewidth}}
\toprule
\textbf{Dataset} & \textbf{Src} & \textbf{Hyp}  \\
\midrule
EPIC & Adopt this working role/setting:\newline Age$=26$; Sex$=$Male; Ethn.$=$Asian; Nat.$=$India; Lang.$=$English.\newline Context: 'We have someone who whenever their car moves uphill (our entire neighbourhood is on a series of huge hills), it lets out an incredibly loud and constant screech. Been like that for months now. Fuck knows what’s wrong with their car but I wish they would fix it at least out of courtesy for the neighbours'\newline Task content: 'Indicate whether this text is ironic or non-ironic: We have one of those boy racers with a big exhaust on his car and everytime he passes our house it sounds like someone is knocking on the door.'. & Answer (final): The text is ironic.\\\midrule
CaLMQA & Respond to this question with your answer: ' Warum wird das Bundesverfassungsgericht als ``Hüter der Verfassung'' genannt?. & This is correct: \newline Schau mal HIER!:\newline    Das Bundesverfassungsgericht in Karlsruhe wacht über die Einhaltung des Grundgesetzes für die Bundesrepublik Deutschland. Seit seiner Gründung im Jahr 1951 hat das Gericht dazu beigetragen, der freiheitlich-demokratischen Grundordnung Ansehen und Wirkung zu verschaffen. Das gilt vor allem für die Durchsetzung der Grundrechte. Zur Beachtung des Grundgesetzes sind alle staatlichen Stellen verpflichtet. Kommt es dabei zum Streit, kann das Bundesverfassungsgericht angerufen werden. Seine Entscheidung ist unanfechtbar. An seine Rechtsprechung sind alle übrigen Staatsorgane gebunden. \newline   Die Arbeit des Bundesverfassungsgerichts hat auch politische Wirkung. Das wird besonders deutlich, wenn das Gericht ein Gesetz für verfassungswidrig erklärt. Das Gericht ist aber kein politisches Organ. Sein Maßstab ist allein das Grundgesetz. Fragen der politischen Zweckmäßigkeit dürfen für das Gericht keine Rolle spielen. Es bestimmt nur den verfassungsrechtlichen Rahmen, innerhalb dessen sich die Politik entfalten kann. Die Begrenzung staatlicher Macht ist ein Kennzeichen des modernen demokratischen Verfassungsstaates.\\
\bottomrule
\end{tabular}
\centering
\caption{Example prompts part 3 (EPIC and CaLMQA)}
\label{tab:example_prompts3}
\end{table*}

\section{Preprocessing Steps \& Data Statistics}
\label{app:further_preprocessing}
As our data stems only from public research datasets, we do not consider its anonymization.
We limit the upper bound of samples from each base dataset to 5000 (before generation) to enable a fair contribution of each dataset. For Mango, we stratify by culture, for BLEnD, GlobalOpinionQA, CulturalBench and Normad, we stratify by country, for INCLUDE we filter for culture and region specific and stratify by country, for EPIC, we stratify by nationality and for CaLMQA by language. For some of these, stratification has no effect if the whole base dataset is selected. The samples from datasets with upper bound are drawn randomly. For the soft errors, the generation is split across 16 processes on 4 Nvidia H100 graphic cards, each running for approximately 2h-4h, amounting to a total of approximately 48 GPU hours for the data generation. For each sample of the base dataset we create up to 2 error samples, one for minor and one for major (if no valid sample is created we do not retry). To build synthetic error reports, we use NLTK word tokenization\footnote{\url{https://www.nltk.org/index.html}} and compute the differences to the original input using the Python \textit{difflib}\footnote{\url{https://docs.python.org/3/library/difflib.html}} library. The number of diff elements is treated as error count. The positions of the differences are the error spans. For the diff processes, we strip all punctuation. We include one correct example for every error sample. Table \ref{tab:data_stats} gives an overview of the dataset statistics. Based on the original datasets' metadata, our final dataset (besides data sourced from Mango) contains content regarding the following regions: 

Afghanistan, Albania, Algeria, Andorra, Angola, Argentina, Armenia, Australia, Austria, Azerbaijan, Bangladesh, Belarus, Belgium, Bolivia, Bosnia and Herzegovina, Brazil, Bulgaria, Burkina Faso, Canada, Chile, China, Colombia, Côte d’Ivoire, Croatia, Cuba, Cyprus, Czechia, Denmark, Ecuador, Egypt, El Salvador, Estonia, Ethiopia, Finland, France, Georgia, Germany, Ghana, Greece, Guatemala, Honduras, Hong Kong (SAR), Hungary, Iceland, India, Indonesia, Iran, Iraq, Ireland, Israel, Italy, Japan, Jordan, Kazakhstan, Kenya, Kuwait, Kyrgyzstan, Laos, Latvia, Lebanon, Libya, Lithuania, Macau (SAR), Malaysia, Maldives, Mali, Malta, Mauritius, Mexico, Mongolia, Montenegro, Morocco, Myanmar, Nepal, Netherlands, New Zealand, Nicaragua, Nigeria, North Korea, North Macedonia, Norway, Pakistan, Palestinian Territories, Peru, Philippines, Poland, Portugal, Puerto Rico, Romania, Russia, Samoa, Saudi Arabia, Senegal, Serbia, Singapore, Slovakia, Slovenia, Somalia, South Africa, South Korea, Spain, Sri Lanka, Sudan, Sweden, Switzerland, Syria, Taiwan, Tajikistan, Tanzania, Thailand, Timor-Leste, Tunisia, Turkey, Uganda, Ukraine, United Kingdom, United States, Uruguay, Uzbekistan, Venezuela, Vietnam, Zimbabwe. 

Additionally, the data sourced from Mango contains 4470 ``cultures'', most of which are regions as well. 

While the dataset is mostly composed of English, it also contains the following languages: Arabic, Chinese, English, German, Hebrew, Hindi, Japanese, Korean, Russian, Spanish, Telugu, Italian, Turkish, Lithuanian, Uzbek, Bulgarian, Armenian, Serbian, Malay, Ukrainian, Belarusian, Polish, Bengali, Azerbaijani, Greek, Dutch; Flemish, Indonesian, Kazakh, North Macedonian, Vietnamese, Finnish, Basque, Tagalog, Croatian, Nepali, Hungarian, Estonian, French, Persian, Tamil, Urdu, Malayalam, Portuguese, Georgian, Albanian.

\section{Baseline Prompts}
\label{app:full_baseline_prompts}
Table \ref{tab:baseline_prompts_full} shows the prompts that we use for baseline evaluation.

\begin{table*}
\begin{tabular}{p{0.15\linewidth} p{0.74\linewidth}}
\toprule
\textbf{Name} & \textbf{Prompt}  \\
\midrule
Counting & Please evaluate the cultural correctness of this instruction and the text that was generated from it.\newline Instruction: '\{Instruction\}'
\newline Text: '\{Text\}'
\newline Please think step-by-step VERY BRIEFLY. Then return the error count. Do not take anything besides cultural correctness into account. If the instruction or the text is culturally correct, the error count should be 0. At the end, return your final count in the format 'COUNT: count'.\\
Severity & Please judge the severity of cultural errors in this input-output pair. \newline
Input: '\{Instruction\}'
\newline Output: '\{Text\}'
\newline Please think step-by-step VERY BRIEFLY. Then return your judgment. Do not take anything besides cultural correctness into account. Return one of the following: none, minor, major, critical
\newline At the end, return your final judgment in the format 'Severity: severity'.\\
Binary & Please judge whether the following input-output pair is culturally problematic.
\newline Input: '\{Instruction\}'
\newline Output: '\{Text\}'
\newline Please think step-by-step VERY BRIEFLY. Then return the judgment. At the end, return your final judgment in the format 'Problem: no problem/very problematic'.\\
\bottomrule
\end{tabular}
\centering
\caption{Baseline Prompts.}
\label{tab:baseline_prompts_full}
\end{table*}

\section{Human annotation}
\label{app:human_annotation} 
We perform human annotation for two different tasks: (1) annotating the cultural relation of samples, the correctness of our synthetically generated errors, including the generated ground truth explanation, error type and error severity and (2) annotating the quality of explanations that were generated by our tuned metrics. In total, 4 annotators annotate 100 samples: 80 English samples annotated by 3 annotators and 8 German, 6 Chinese and 6 Japanese samples are annotated by 2 annotators. The non-English samples stem from CaLMQA. Each data sample is annotated by at maximum one co-author. The other annotators are a student assistant and one annotator from Upwork (\url{https://www.upwork.com/}). Overall, the annotation time was approx.\ 4 h - 6 h per person (one person only annotated multilingual examples) at an estimated cost of 340€. 

\begin{figure*}[!ht]
    \centering
    \includegraphics[width=0.95\linewidth]{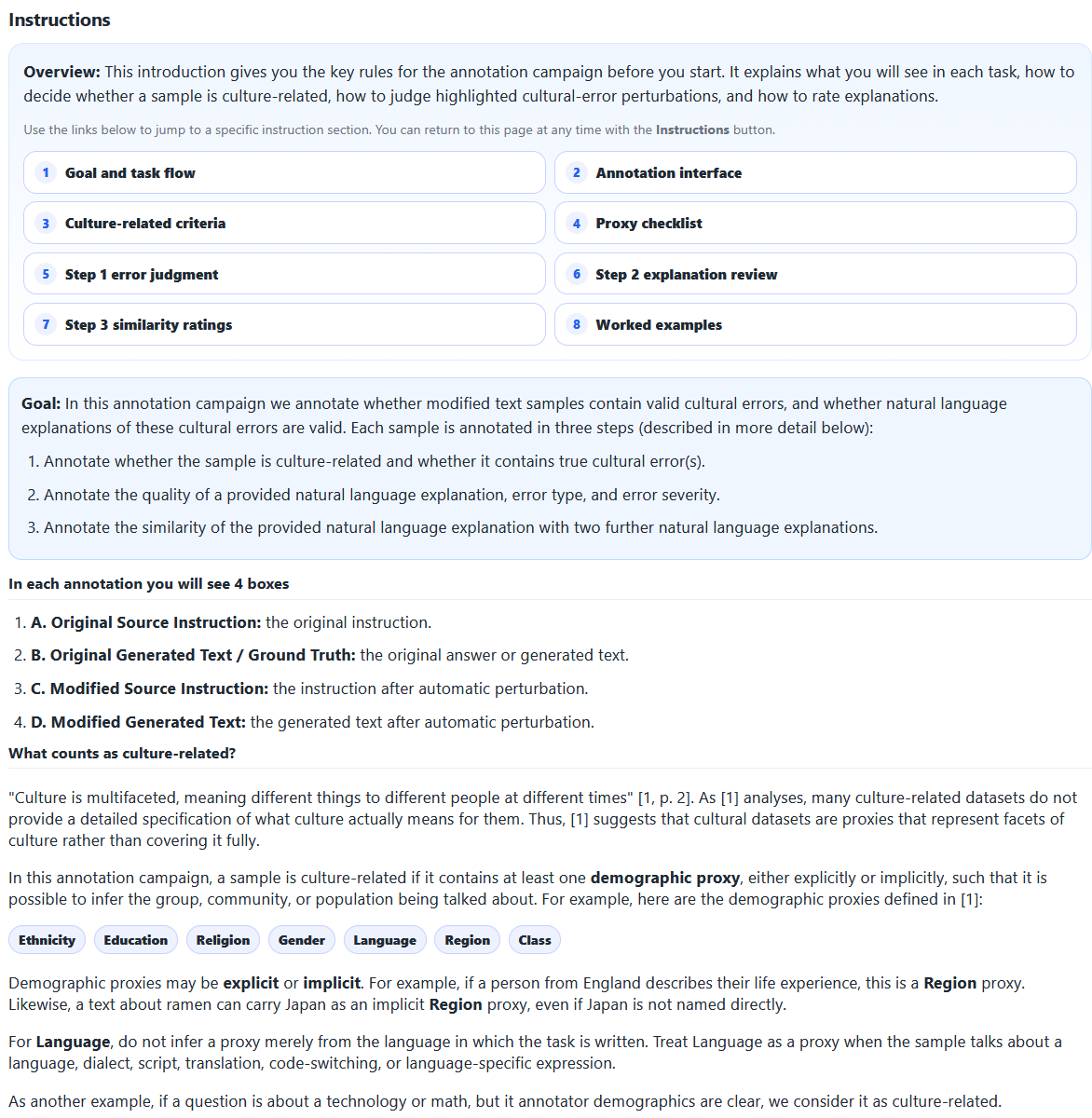}
    \caption{A screenshot of the instructions displayed in our annotation interface. (Part 1)}
    \label{fig:annot1}
\end{figure*}

\begin{figure*}[!ht]
    \centering
    \includegraphics[width=0.95\linewidth]{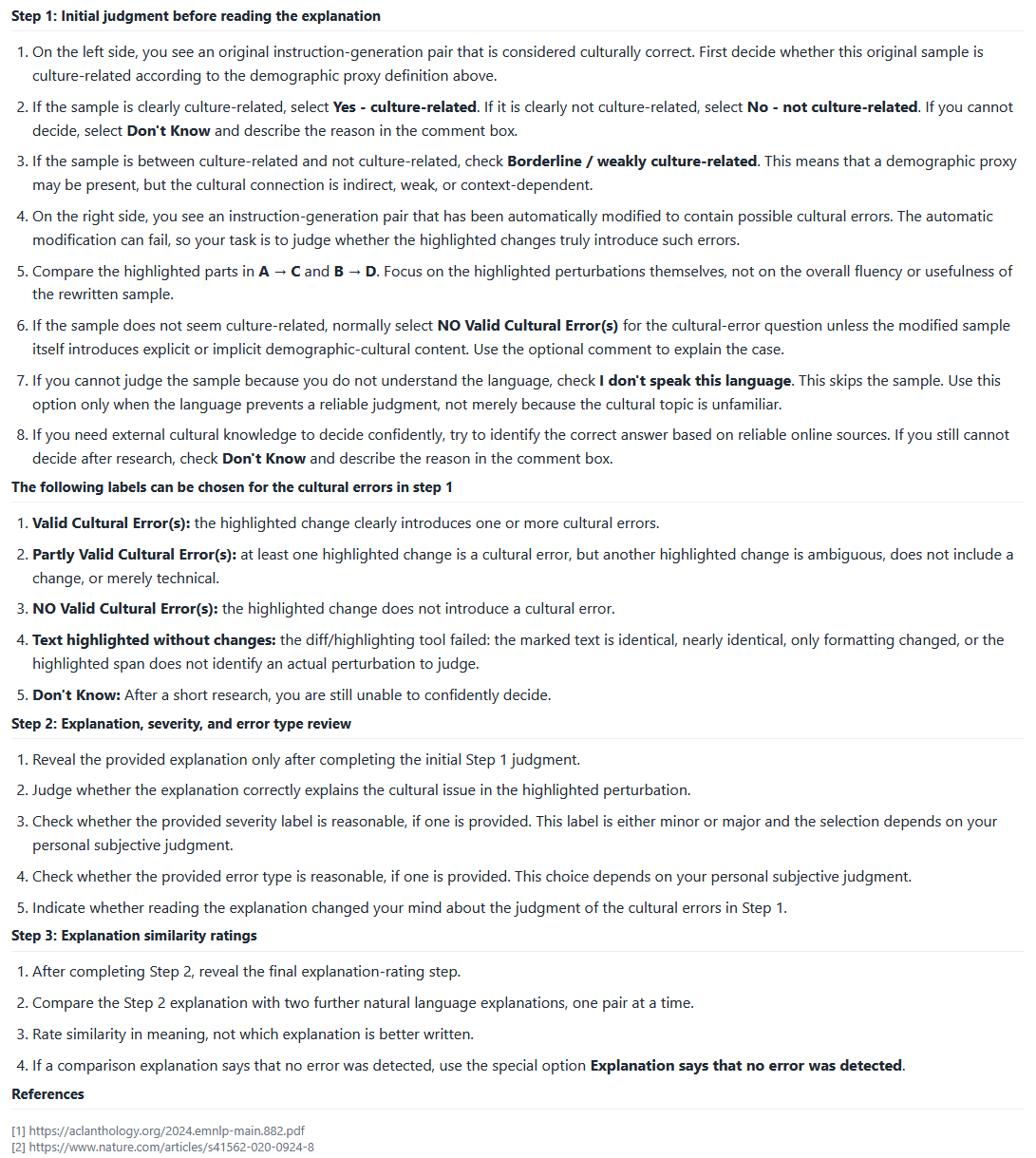}
    \caption{A screenshot of the instructions displayed in our annotation interface. (Part 2)}
    \label{fig:annot2}
\end{figure*}

\begin{figure*}[!ht]
    \centering
    \includegraphics[width=0.95\linewidth]{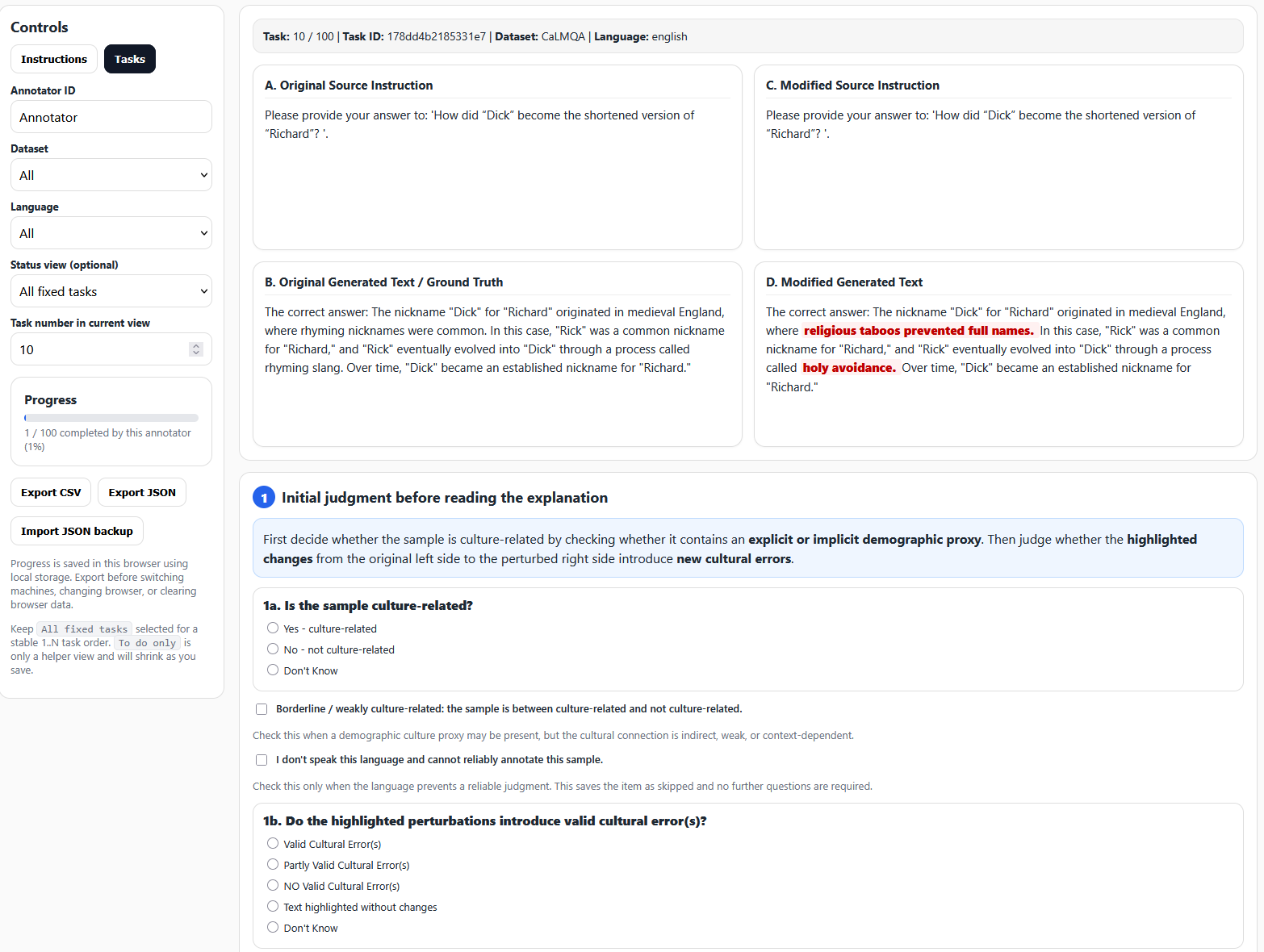}
    \caption{A screenshot of our annotation interface. (Step 1)}
    \label{fig:annot3}
\end{figure*}

\begin{figure*}[!ht]
    \centering
    \includegraphics[width=0.95\linewidth]{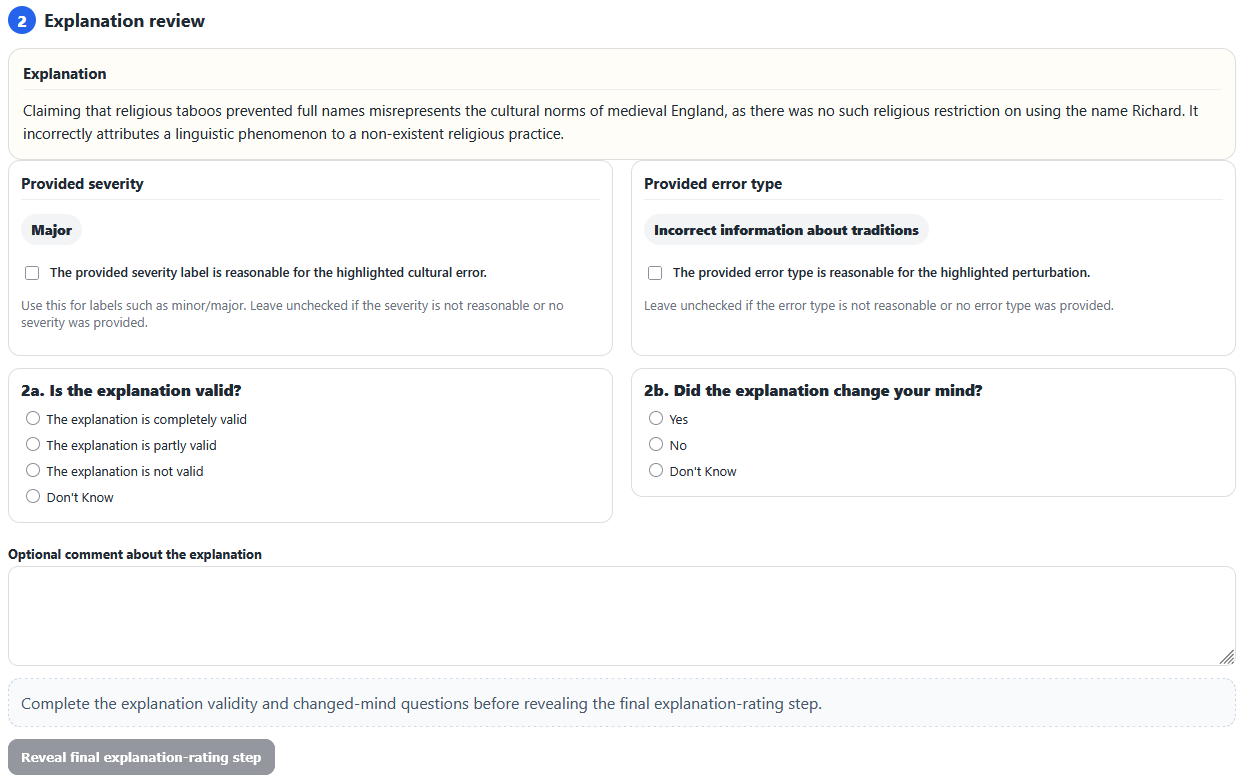}
    \caption{A screenshot of our annotation interface. (Step 2)}
    \label{fig:annot4}
\end{figure*}

\begin{figure*}[!ht]
    \centering
    \includegraphics[width=0.95\linewidth]{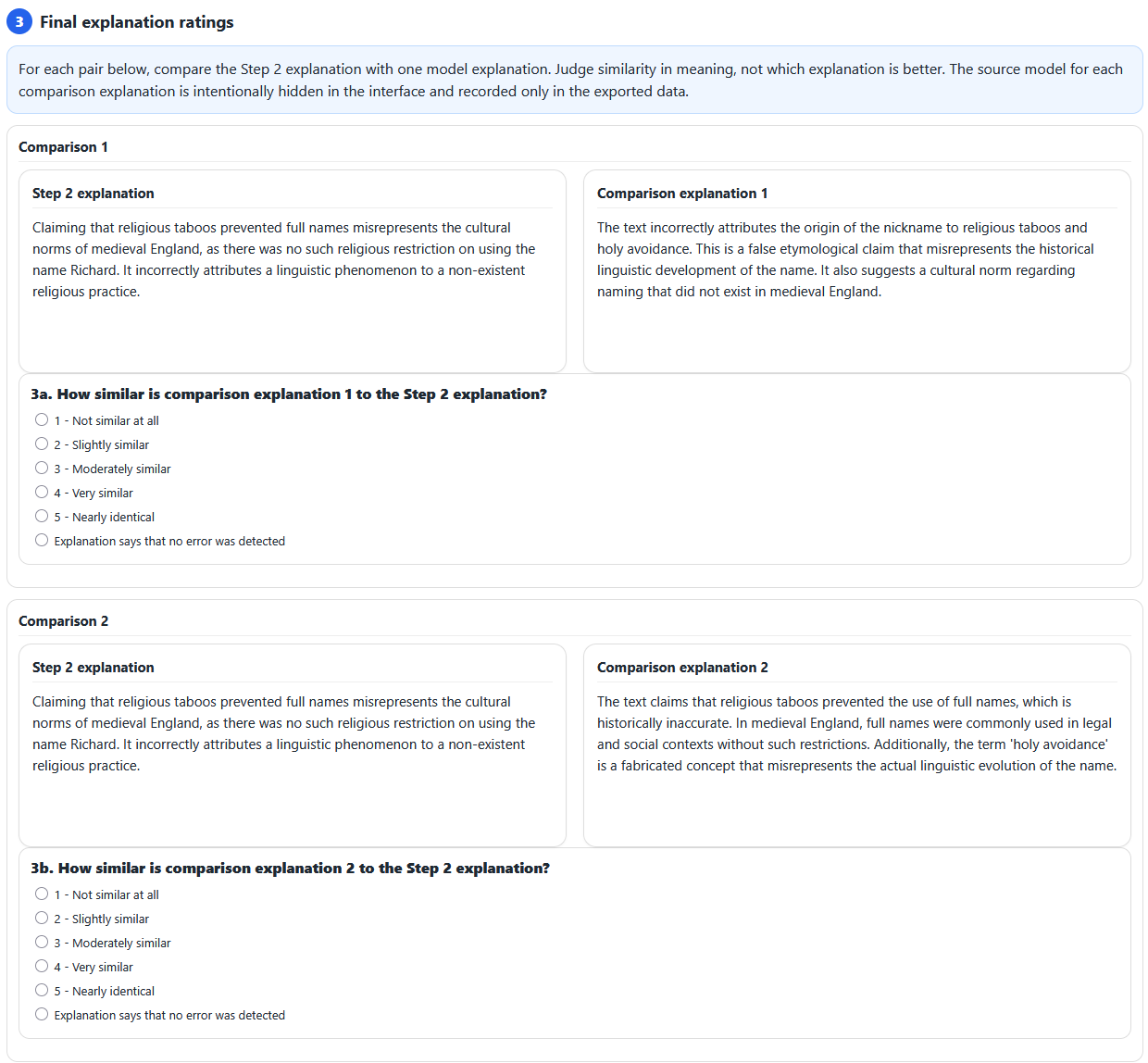}
    \caption{A screenshot of our annotation interface.  (Step 3)}
    \label{fig:annot5}
\end{figure*}

Each sample is annotated in a three-step process: first the cultural relation and correctness of the error is annotated, then the explanation details and lastly the quality of the metric explanations. 
We provide detailed annotation instructions to disambiguate the different cases. Notably, the \textit{error severity} is treated as a subjective category, as we want to see if the severity impression aligns with the ones provided by the models. Figure \ref{fig:annot1}, Figure \ref{fig:annot2}, Figure \ref{fig:annot3}, Figure \ref{fig:annot4} and Figure \ref{fig:annot5}  show sample screenshots of our annotation interface. Additionally, participants were informed that their data is collected for a study on cultural errors. Some of the 100 samples contain languages that are not spoken by all of the annotators (to enable multilingual evaluation). In these cases, the annotators can skip these samples. This includes a maximum of 12 cases for one annotator. Additionally, if an annotator is not familiar with a culture, they are asked to double check the introduced errors with reliable online sources. As the ground truth data from the original datasets is provided, judging the cultural errors is simple in many cases.


Table \ref{tab:multilingual-majority-votes} shows the percentage of samples per language where both annotators agree on positive classes. It shows that high-resource languages besides English also achieve good quality. Only the explanation quality is less agreed on with 50\% for Chinese and German. 

\begin{table}[t]
\begin{tabular}{lrrr}
\toprule
Criterion & Japanese & Chinese & German \\
\midrule
N & 6 & 6 & 8 \\
Culture rel. & 83.3 & 83.3 & 100.0 \\
Error valid  & 83.3 & 66.7 & 100.0 \\
Expl. valid  & 83.3 & 50.0 & 50.0 \\
\bottomrule
\end{tabular}
\caption{Majority vote results for positive non-english classes. Error valid counts partially correct span annotations as complete errors.}
\label{tab:multilingual-majority-votes}
\end{table}

Table \ref{tab:auxiliary-majority-all-languages} shows the rate with which annotators agree on severity labels (minor vs.\ major), the returned error type and how often the explanation convinced them of a cultural error. It shows that the minor severity is much less certain (at 51\% agreed in majority voting), compared to major severity (at 96\%). This means that cultural errors tend to be perceived as stronger than minor. Also, 4\% of explanations convinced the annotators of a different choice. 

\begin{table}[t]
\centering
\begin{tabular}{lrrr}
\toprule
Criterion & Yes & N & \% \\
\midrule
Severity OK (minor) & 31 & 60 & 51.7 \\
Severity OK (major) & 38 & 40 & 95.0 \\
Type OK & 96 & 100 & 96.0 \\
Changed mind & 4 & 100 & 4.0 \\
\bottomrule
\end{tabular}
\caption{Auxiliary annotation results across all languages using task-level majority votes. With two annotations, both annotators must vote yes for a positive result.}
\label{tab:auxiliary-majority-all-languages}
\end{table}

\section{License}
\label{sec:license}
We use pre-trained models including Qwen3.5-122B-A10B, Qwen3-14B and Mistral-24B-Instruct-2501 under the \textit{Apache-2.0 License}, Phi-3-mini-4k-Instruct, Phi-4, and DeepSeek-R1-Distill-Qwen-14B under the \textit{MIT License}, Gemma-3-27B-it under the \textit{Gemma Terms of Use}. 

As source datasets, we use BLEnD under the \textit{Creative Commons Attribution Share Alike 4.0 International (CC BY-SA 4.0) License}, EPIC, NativQA and GlobalOpinionQA under the \textit{Creative Commons Attribution Non Commercial Share Alike 4.0 International (cc-by-nc-sa-4.0) License}, INCLUDE under the \textit{Apache License 2.0}, CulturalBench, Mango and Normad under the \textit{Creative Commons Attribution 4.0 International (CC BY 4.0) License}, and CaLMQA under the \textit{MIT License}.
All usage is consistent with the intended research purposes of these resources.

\section{Scaled accuracy across languages}
\label{app:lang}
\begin{figure*}[!ht]
    \centering
    \includegraphics[width=0.95\linewidth]{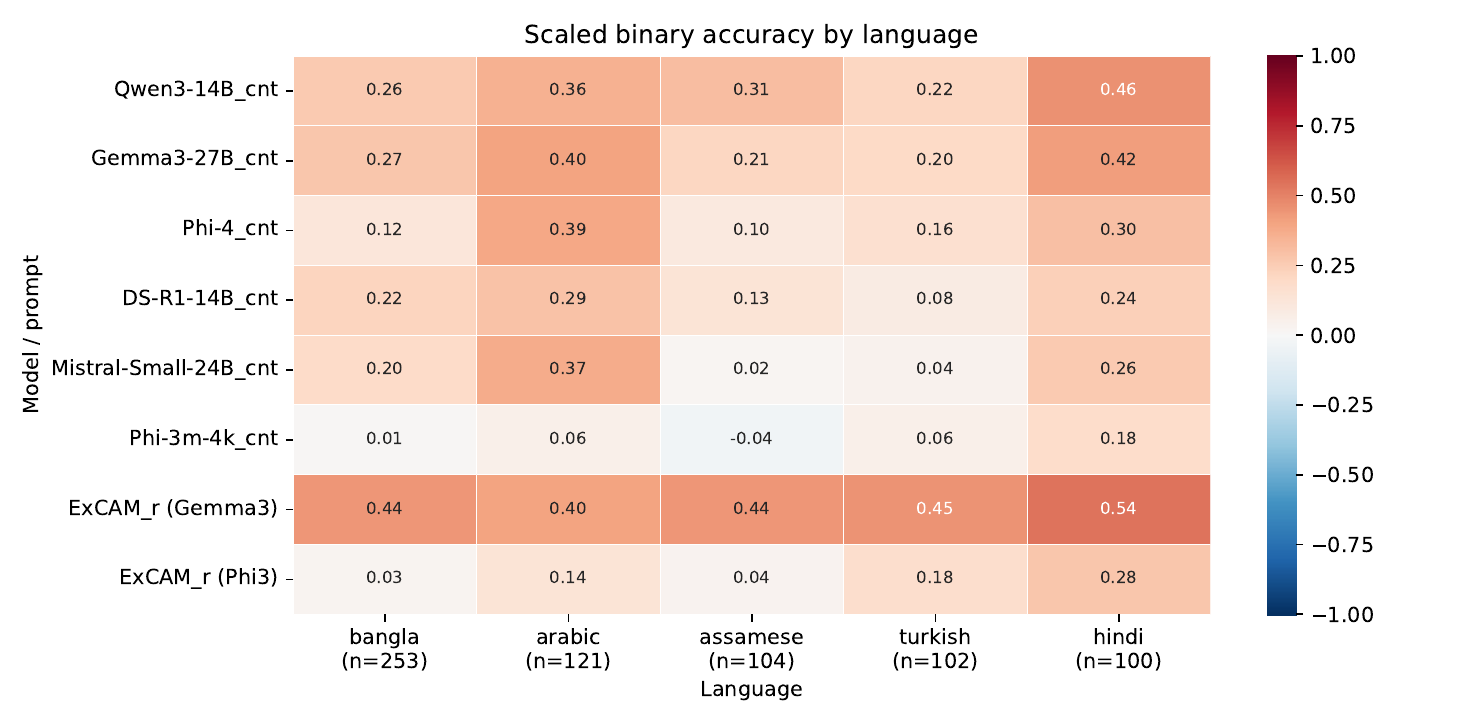}
    \caption{Scaled accuracy across 5 languages.}
    \label{fig:langs}
\end{figure*}

As an ablation, in Figure \ref{fig:langs}, we report the scaled accuracy for 5 of the languages that have more than 100 samples in the dataset. Interestingly, the performance difference is 0.14.  We can see that the Gemma-based ExCAM outperforms the other models across languages, while the Phi base-model and tuned model have a weaker performance. 

\section{Scaled accuracy across all LoRAs}
\label{app:all_loras}
\begin{figure*}[!ht]
    \centering
    \includegraphics[width=0.95\linewidth]{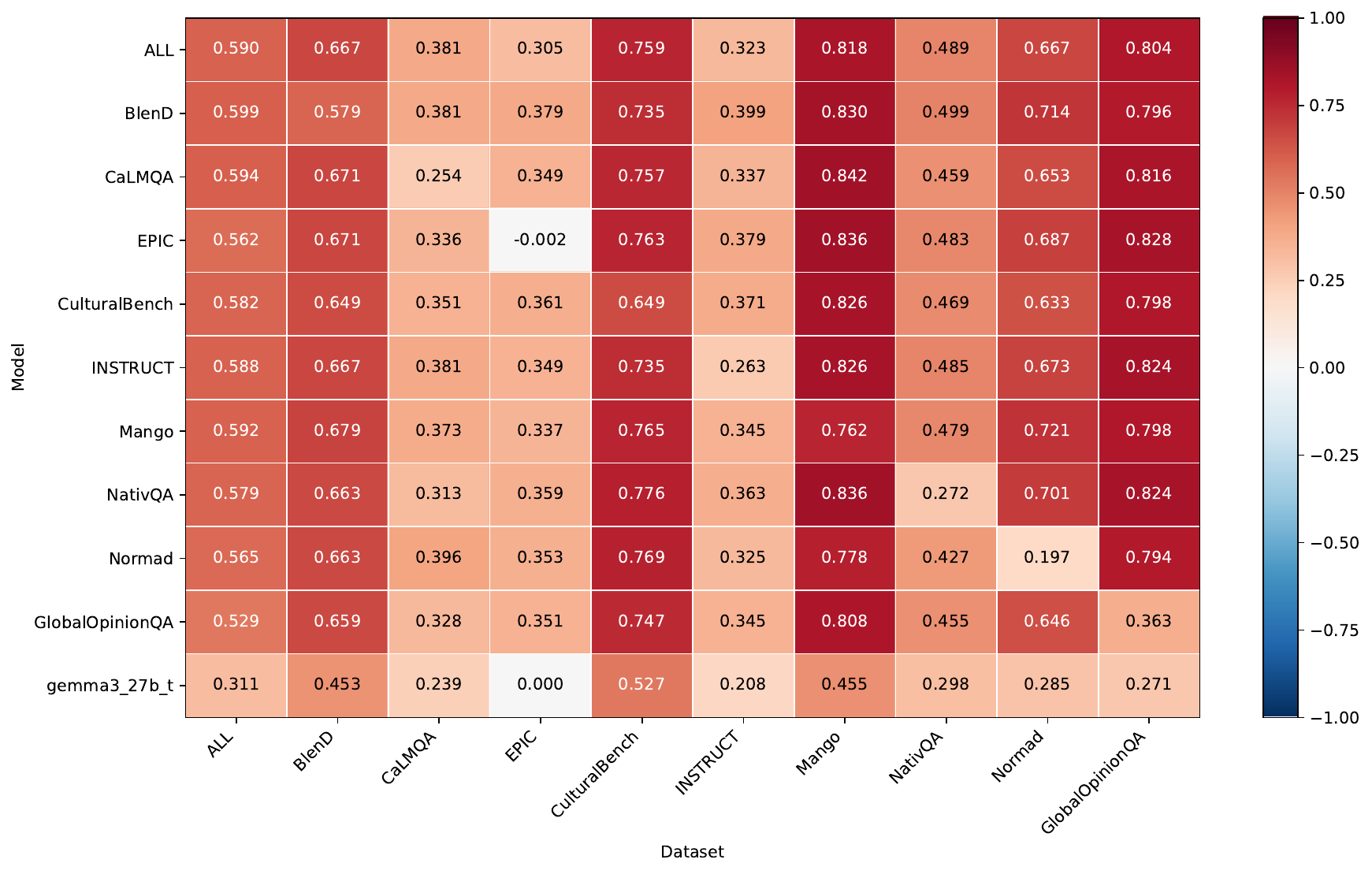}
    \caption{Scaled accuracy of leave-one-out trained LoRAs on all sub datasets.}
    \label{fig:lorasall}
\end{figure*}

Figure \ref{fig:lorasall} shows the performance of all LoRAs on all datasets. This shows that the in-domain performance is only weakly influenced by one out-of-domain LoRA being removed from the training data.

\end{document}